\documentclass[runningheads]{llncs}
\usepackage{pifont}
\newcommand{\cmark}{\ding{51}}%
\newcommand{\xmark}{\ding{55}}%
\usepackage[T1]{fontenc}
\usepackage{amsmath}
\usepackage{amssymb}
\usepackage{textcomp}
\usepackage{booktabs}
\usepackage{subfig}
\usepackage{caption}
\usepackage[section]{placeins}

\usepackage{multirow}
\usepackage{multicol}
\usepackage{graphicx}
\usepackage{comment}
\usepackage{ragged2e}
\usepackage{microtype}
\begin{document}

\title{SYNPRED: A Synergistic Approach to Multimodal Learning for Clinical Prediction}

\author{Ivana Janíčková\inst{1,2,5}\and
Yen Y. Tan \inst{3}\and
Thomas H. Helbich\inst{6}\and
Philipp Seeböck \inst{1,2}\and
Julieta Di Marco \inst{1,2}\and
Zsuzsanna Bago-Horvath\inst{4}\and
Ulrike Heber\inst{4}\and
Thomas Spiegel \inst{6}\and
Georg Langs\inst{1,2,5}
}  
\authorrunning{I. Janíčková et al.}
\institute{Computational Imaging Research Lab, Department of Biomedical Imaging and Image-guided Therapy, Medical University of Vienna, Austria 
\and 
Comprehensive Center for Artificial Intelligence in Medicine, Medical University of Vienna, Austria \and
Department of Obstetrics and Gynecology, Medical University of Vienna, Austria \and
Department of Pathology, Medical University of Vienna, Austria \and
Christian Doppler Laboratory for Machine Learning Driven Precision Imaging, Department of Biomedical Imaging and Image-guided Therapy, Medical University of Vienna, Austria\and
Division of General and Pediactric Radiology, Department of Biomedical Imaging and Image-guided Therapy, Medical University of Vienna, Austria\\
\email{ivana.janickova@meduniwien.ac.at}, 
\email{georg.langs@meduniwien.ac.at} 
\texttt{\url{https://www.cir.meduniwien.ac.at}}
}
\titlerunning{Synergistic Multimodal Learning}
\maketitle              

\begin{abstract}
Multimodal medical data captures complementary aspects of disease. It provides a comprehensive basis for diagnosis and the identification of biological associations linked to progression or treatment response. Established contrastive learning approaches exploit shared information across modalities, but fall short in exploiting complementary, modality-specific signals relevant for a task. Here, we propose a synergistic multimodal variational autoencoder (VAE) framework that separates task-relevant unimodal and multimodal information from instance-specific variation task-irrelevant, allowing complementary signals across modalities to be leveraged for prediction. The method yields benefits even if modalities are missing at test time. We evaluate the method on a treatment response prediction task using medical imaging and RNA data. Empirical comparison with alignment-based and multimodal VAE state-of-the-art approaches demonstrates improved multimodal prediction performance, and consistent benefits for prediction from a single modality after multimodal training. The full code
can be found here: \url{https://github.com/cirmuw/SYNPRED}. 
\end{abstract}


%
%
%
\section{Introduction}
Multimodal data can improve predictive performance and clinical decision support in medical machine learning \cite{huang2020fusion}. Importantly, it often combines modalities describing complementary aspects of disease, sharing only limited semantic overlap. For instance, radiological imaging may reflect tumour structure, whereas biopsy-derived molecular data characterises underlying biological processes such as methylation or gene expression \cite{guan2025mvnmf,kawahara2025integrating}. Multimodal data integration approaches based on alignment (contrastive or attention-based) are fundamentally limited, as they primarily capture shared information and fail to explicitly model complementary, modality-specific signals that nevertheless might be informative for the prediction task \cite{dufumier2025align}.

Here we propose a multimodal generative framework that learns structured latent representations for clinical prediction. Instead of representing only characteristics shared across modalities, our model transparently disentangles (1) shared and (2) complementary (modality-specific) task-relevant features, and (3) residual features independent of the prediction target.  
We show that explicitly modeling shared and complementary task-relevant features enables the model to capture synergistic cross-modal information —information that cannot be recovered from any single modality alone \cite{dufumier2025align,shi2019variational} -  during multimodal training, improving downstream multimodal and single-modality prediction.

\paragraph{Related Work.}
Multimodal representation learning includes variational approaches for shared representation learning and handling of missing modalities \cite{lee2021private,shi2019variational,sutter2024unity}, as well as alignment-based contrastive and attention-based methods \cite{radford2021learning,jia2021scaling,chen2020uniter,lu2019vilbert}. In medical applications, most methods focus on data with strong semantic overlap and rely on contrastive alignment or attention mechanisms \cite{wang2022medclip,yan2025pathway,huang2021gloria,chen2021multimodal}. As noted by \cite{dufumier2025align}, such approaches primarily emphasize shared information and may under represent complementary or modality-specific task-relevant signals, particularly when semantic correspondence between modalities is limited. Classical radiomics and radiogenomics methods address multimodal data with lower semantic overlap via feature correlation \cite{guan2025mvnmf,kawahara2025integrating}. Deep learning approaches typically rely on late fusion through feature concatenation \cite{tripathi2022radgennets,huang2023deep}. Only a small number of recent contributions propose more structured solutions for such data, including alignment-based \cite{taleb2022contig} and attention-based approaches \cite{ai2024sama,li2024mri}. Despite their differing mechanisms, these methods remain limited in their ability to handle missing modalities and to explicitly model complementary, modality-specific task information.  Recent analyses highlight the importance of such complementary information, 
that multimodal pretraining can uncover \textit{synergistic} cross-modal patterns beyond shared information. While synergy arises from multimodal interactions, learning these patterns during training can enhance unimodal representations and improve single-modality performance in downstream tasks \cite{shi2019variational,dufumier2025align}.

\paragraph{Contribution.}
We propose SYNPRED, a synergistic multimodal VAE for representation learning in radiology-genomics settings with limited semantic overlap between modalities. Unlike alignment-based approaches that primarily emphasize shared information, our framework explicitly disentangles shared task-relevant (\textit{fusion}), modality-specific task-relevant (\textit{uni}), and task-irrelevant instance-specific (\textit{inst}) latent components, enabling complementary signals to contribute to prediction. We evaluate the framework on the public BreastDCEDL dataset \cite{fridman2026breastdcedl}, combining paired breast MRI and gene expression data from the I-SPY2 cohort \cite{li2022spy,newitt2025acrin}. Experiments show improved multimodal performance and consistent gains in single-modality prediction, particularly for imaging, providing empirical evidence that multimodal training improves representation learning through synergistic cross-modal interactions.

\section{Method}

\begin{figure}[t]
    \centering
    \includegraphics[width=1.0\linewidth]{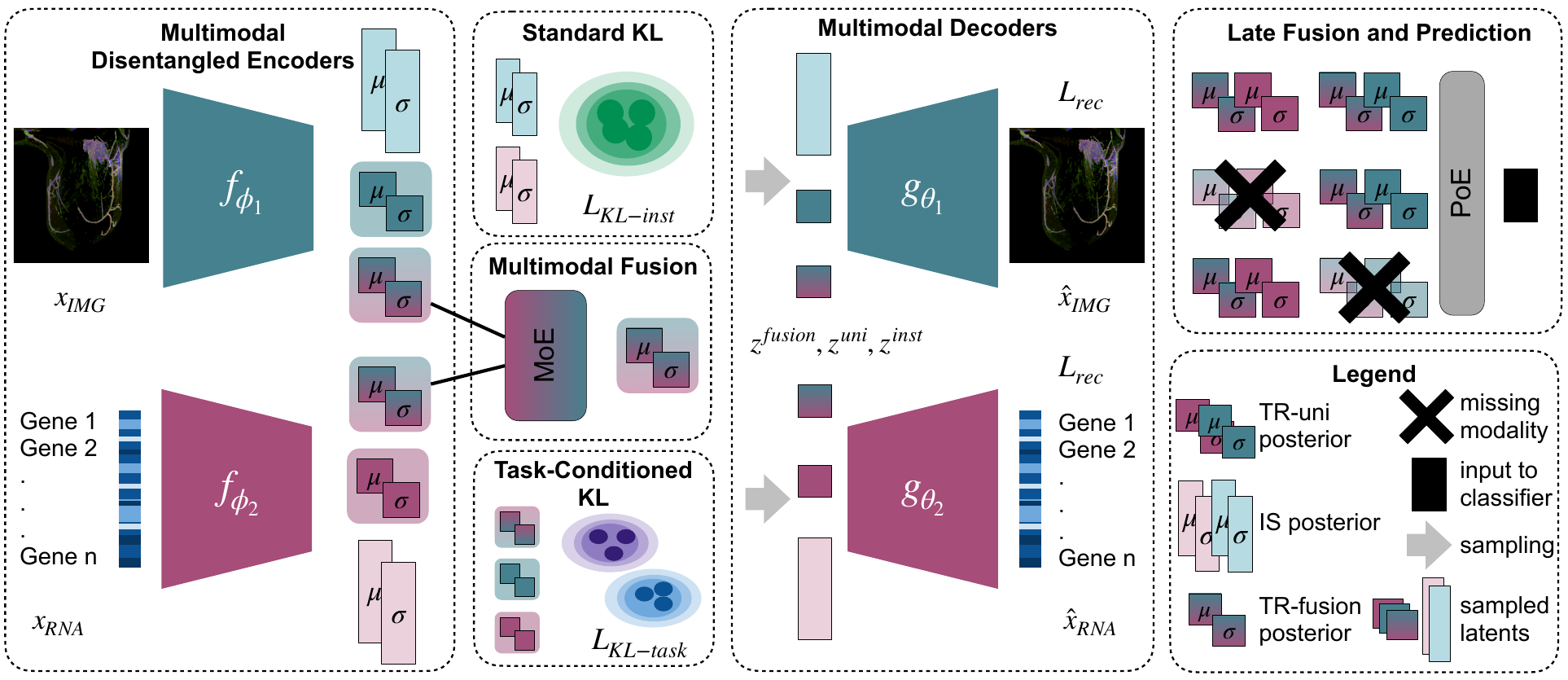}
    \caption{\textbf{Method Overview} Each modality-specific encoder maps the input to a factorised Gaussian posterior, decomposing the latent space into shared task-relevant (TR-fusion), modality-specific task-relevant (TR-uni), and instance-specific (IS) components with posterior parameters ($\mu,\sigma$) and sampled latents ($z$). TR-fusion posteriors of each modality are combined via a mixture-of-experts (MoE). Task-conditioned KL regularisation ($\mathcal{L}_{\text{KL-task}}$) is applied to TR-fusion and TR-uni posteriors, while a KL term ($\mathcal{L}_{\text{KL-inst}}$) regularises IS posteriors. The sampled latents $z$ are concatenated and decoded using a reconstruction loss ($\mathcal{L}_{\text{rec}}$). For prediction, embeddings are combined using a product-of-experts (PoE).}
    \label{fig:main}
\end{figure}
We propose \textbf{SYNPRED}, a \textbf{syn}ergistic \textbf{pred}iction model for learning disentangled, task-relevant representations from medical modalities using a multimodal VAE framework. As illustrated in Fig.~\ref{fig:main}, each modality is modeled using multimodal task-relevant information (\textit{fusion}), modality-specific task-relevant information (\textit{uni}), and instance-specific variation (\textit{inst}). The model is trained with reconstruction and variational regularization under standard and task-conditioned priors, enabling robust prediction under missing modalities via mixture- and product-of-expert fusion. The following problem formulation formalizes this structure.

\subsubsection{Problem Formulation}
We consider a two-modality setting with inputs $X_1, X_2$ and a task label $Y$. The goal is to predict $Y$ from multimodal observations by exploiting shared and complementary information across modalities. Motivated by the decomposition of multimodal information into redundant and complementary components \cite{bertschinger2014quantifying}, we assume that each modality is generated from instance-specific variation, modality-specific task-relevant information, and shared task-relevant information. We model this structure using disentangled latent representations that preserve complementary and shared signals, enabling synergistic task-relevant information to emerge.

\subsection{SYNPRED: Synergistic Multimodal VAE}

\subsubsection{Latent Decomposition and Reconstruction}
We instantiate the framework for images and gene expression data using one encoder--decoder pair per modality and a shared latent space. Each encoder $f_{\phi_m}$ maps modality input $x_m$ to factorised Gaussian posteriors over shared task-relevant (TR-fusion), modality-specific task-relevant (TR-uni), and instance-specific (IS) latent components:
\begin{equation}
q_{\phi_m}(z_m^{\text{fusion}}, z_m^{\text{uni}}, z_m^{\text{inst}} \mid x_m)
=
\underbrace{
q_{\phi_m}(z_m^{\text{fusion}} \mid x_m)
}_{\text{TR-fusion}}
\underbrace{
q_{\phi_m}(z_m^{\text{uni}} \mid x_m)
}_{\text{TR-uni}}
\;
\underbrace{
q_{\phi_m}(z_m^{\text{inst}} \mid x_m)
}_{\text{IS}}
.
\end{equation}
The total dimensionality is $d = 2d_t + d_i$, where $d_t$ and $d_i$ denote the dimensionalities of task-relevant and instance-specific components, respectively. For each latent component $z_m^p$, the encoder outputs the mean and variance of the approximate posterior
\begin{equation} 
\label{eq:posterior}
q_{\phi_m}(z_m^{p} \mid x_m) = 
\mathcal{N}\!\left(
\mu_m^{p}(x_m),
\mathrm{diag}(\sigma_m^{p}(x_m)^2)
\right),
\qquad
 p \in \{\text{fusion},\text{uni},\text{inst}\}.
\end{equation}

The decoder $g_{\theta_m}$ reconstructs
$
\hat{x}_m = g_{\theta_m}(z_m^{\text{fusion}}, z_m^{\text{uni}}, z_m^{\text{inst}})
$
for $m \in \{1,2\}$ from the concatenated sampled latents. Reconstructions are optimised using an $\ell_1$ loss, while instance-specific latents (IS) are regularised toward a standard Normal prior using the Kullback--Leibler (KL) divergence:
\begin{equation}
\mathcal{L}_{\text{rec}}^{(m)} =
\mathbb{E}\!\left[\lVert \hat{x}_m - x_m \rVert_1\right],
\qquad
\mathcal{L}_{\text{KL-inst}}^{(m)} =
\mathrm{KL}\!\left(
q_{\phi_m}(z_m^{\text{inst}} \mid x_m)
\;\|\;
\mathcal{N}(0,I)
\right).
\end{equation}

\subsubsection{Multimodal Fusion}
Fusion task-relevant posteriors (TR-fusion) of each modality $q_{\phi_m}(z_m^{\text{fusion}} \mid x_m)$ are combined using mixture-of-experts (MoE) \cite{shi2019variational} over the set of available modalities $\mathcal{M}$.
The parameters of multimodal fused posterior $q(z^{\text{fusion}} \mid x_1,x_2)$ are computed as weighted averages of the modality-specific posteriors, with learnable mixture weights $w_{m}$ for each modality:
\begin{equation}
\mu^{\text{fusion}} =
\sum_{m \in \mathcal{M}} w_m \mu_m^{\text{fusion}},
\qquad
\sigma^{\text{fusion}} =
\sum_{m \in \mathcal{M}} w_m \sigma_m^{\text{fusion}}.
\end{equation}

\subsubsection{Task Conditioning}

The task-relevant posteriors (TR-fusion, TR-uni) $q_{\phi_m}(z_m^{p} \mid x_m)$, $p \in \{\text{fusion}, \text{uni}\}$, are shaped using an observed target label $y$. Inspired by \cite{kingma2014semi}, we condition the generative model on $y$ by defining task-conditional Gaussian priors:
$
p(z^{p} \mid y) = \mathcal{N}\!\left(
\mu^{p}_y,\;
\mathrm{diag}(\sigma^{p2}_y)
\right),
$
which are used to construct the task-conditioned KL loss $\mathcal{L}_{\text{KL-task}}$. Here, separate parameters $\{\mu^{p}_y, \sigma^{p}_y\}$ are learned for each latent type $p$ and target label $y$, encouraging the $z^{\text{fusion}}$ latents to capture shared task-related information and the $z^{\text{uni}}$ latents to capture modality-specific complementary task-related information. Task conditioning is enforced via KL divergence terms between the approximate posteriors and the corresponding task-conditional priors:
\begin{equation}
\begin{aligned}
\mathcal{L}_{\text{KL-task}}
={}&
\mathrm{KL}\!\left(
q(z^{\text{fusion}} \mid x_1, x_2)
\;\|\;
\mathcal{N}\!\left(
\mu^{\text{fusion}}_y,\;
\mathrm{diag}(\sigma^{\text{fusion}}_y{}^{2})
\right)
\right)
\\
&+
\mathrm{KL}\!\left(
q_{\phi_m}(z_m^{\text{uni}} \mid x_m)
\;\|\;
\mathcal{N}\!\left(
\mu^{\text{uni}}_y,\;
\mathrm{diag}(\sigma^{\text{uni}}_y{}^{2})
\right)
\right).
\end{aligned}
\end{equation}

\subsubsection{Overall Loss}

The model is trained by minimizing a weighted sum of reconstruction and the two KL-based terms with $\beta$ is a fixed hyperparameter controlling the strength of latent regularisation:
\begin{equation}
\mathcal{L_\text{total}}
=
\mathcal{L}_{\text{rec}}
+
\beta
\left(
\mathcal{L}_{\text{KL-inst}}
+
\mathcal{L}_{\text{KL-task}}
\right),
\label{eq:total_loss}
\end{equation}

\subsubsection{Prediction}
For downstream prediction, embeddings are formed by joining posterior parameters of the shared task-relevant (TR-fusion) and modality-specific task-relevant (TR-uni) latents,
$
(\mu_m,\log\sigma_m^2)=([\mu_m^{\text{fusion}};\mu_m^{\text{uni}}],
[\log\sigma_m^{2,\text{fusion}};\log\sigma_m^{2,\text{uni}}]).
$
Modality-specific Gaussian experts are combined via a product-of-experts (PoE), and the fused mean is used as the classifier input. Representations are evaluated via linear probing and fine-tuning, trained on multimodal embeddings and tested on unimodal or multimodal inputs without retraining.
\section{Experiments and Results}
\subsubsection{Data}
We use the multi-center public BreastDCEDL dataset \cite{fridman2026breastdcedl}, which contains pretreatment breast cancer (BC) patients with a binary pathological complete response (pCR) label (1452 patients). The dataset aggregates dynamic contrast-enhanced (DCE) MRI from the I-SPY1 \cite{newitt9multi}, I-SPY2 \cite{li2022spy,newitt2025acrin}, and Duke \cite{saha2021dynamic} cohorts, and we adopt the official train/validation/test split.
For multimodal experiments, the I-SPY2 cohort (982 patients) additionally provides paired pretreatment gene-expression profiles (GEO: GSE194040) \cite{petricoin2025immune,wolf2022redefining}, containing measurements for over 19000 genes. From the DCE MRI, we compute a 2D maximum-intensity projection of early and late subtraction images. Images are intensity-normalised to $[0,1]$. RNA features are mean-imputed, log-transformed, and z-score normalised. During training, we apply simple noise-based augmentation, using Gaussian and Gibbs noise for images and Gaussian noise for RNA inputs. 
\subsubsection{Implementation Details}
The model consists of two modality-specific VAE branches and a fusion module. The image branch consists of convolutional, and the RNA branch of fully-connected layers with hidden size 256. Latent variables are partitioned into \textit{fusion, uni} and \textit{inst} components, with total dimensionality $d=256$ ($d_c=16$ per task-related component). Multimodal fusion is performed via MoE over modality-specific fusion posteriors, with learnable softmax weights. The model is trained using separate learning rates for the image and RNA branches ($10^{-5}$ and $10^{-6}$, respectively) and a fixed regularisation weight $\beta=10^{-4}$ in Eq.  \ref{eq:total_loss}. Models are trained for 300 epochs (batch size 16) across five seeds, with selection based on validation AUROC. Ablations alter latent disentanglement and associated losses. Performance is evaluated via linear probing and MLP fine-tuning, with missing RNA handled through MoE and PoE fusion.

\subsection{Results}
\begin{table}[!t]
    \centering
    \caption{\textbf{Comparison with SOTA:} Mean AUROC and 95\% confidence interval results for image/RNA-only, and multimodal evaluation.$\dag$ and $\ddag$ indicate complete modality requirement during pretraining and fine-tuning respectively. All baselines were finetuned, for SYNPRED (lin.prob.) is linear probing; (ft) is finetuning.}
    \label{tab:results}
    {\fontsize{8}{10}\selectfont
    \begin{tabular}{lccc}
    \toprule
    \multirow{2}{*}{Method} & IMG-only & RNA-only & Multimodal \\
                           & AUROC   & AUROC    & AUROC      \\
    \midrule
    
    \multicolumn{4}{l}{\hspace{0mm}\textit{\textbf{Single-Modal Approaches}}}  \\
    \midrule
   
    IMG VAE \hfill &
    0.549 [0.508,0.590] & --- & --- \\

    RNA VAE \hfill &
    --- & 0.656 [0.614,0.702] & --- \\

    \midrule
    
    \multicolumn{4}{l}{\hspace{0mm}\textit{\textbf{Multimodal Approaches}}}  \\
    \midrule
    
    concat \hfill $\ddag$ &
    --- & --- & 0.644 [0.630,0.657] \\
    
    ContIG \cite{taleb2022contig} \hfill $\dag$  &
    0.560 [0.553,0.566] &
    0.702 [0.682,0.715] &
    0.696 [0.681,0.709] \\
    
    MMVM VAE \cite{sutter2024unity} \hfill $\dag$ &
    0.512 [0.456,0.566] &
    0.634 [0.604,0.659] &
    0.628 [0.594,0.661] \\
    
    DMVAE \cite{lee2021private} \hfill &
    0.542 [0.534,0.550] &
    0.668 [0.655,0.683] &
    0.670 [0.653,0.685] \\   
    \midrule
        
    SYNPRED (lin.prob.) &
    \underline{0.638 [0.601,0.673]} &
    \underline{0.706 [0.666,0.741]} &
    \textbf{0.725 [0.682,0.757]} \\
    
    SYNPRED (ft) &
    \textbf{0.693 [0.620,0.740]} &
    \textbf{0.717 [0.698,0.734]} &
    \underline{0.707 [0.695,0.719]} \\

    \bottomrule
    \end{tabular}
    }
\end{table}
\subsubsection{Comparison with state-of-the-art approaches}
SYNPRED outperforms state-of-the-art (SOTA) methods in multimodal prediction of treatment response (pCR) for BC patients. Even if only a single modality is available during test time, it substantially outperforms alternative approaches. This is particularly pronounced for the prediction from imaging data where simple uni-modal approach yields AUROC of $0.549$ while SYNPRED achieves and AUROC of $0.693$. This suggests that RNA data can help the model to better identify predictive signatures in the imaging data.  We compare against several pretraining approaches fine-tuned with an MLP classifier (Table  \ref{tab:results}), including unimodal VAEs and a simple concatenation-based multimodal baseline. For alignment-based pretraining, we use ContIG \cite{taleb2022contig}, a contrastive framework that enforces cross-modal representation alignment. As multimodal VAE baselines, we include MMVM VAE \cite{sutter2024unity}, which uses a data-dependent prior for soft multimodal information sharing, and DMVAE \cite{lee2021private}, which disentangles multimodal - neither method explicitly disentangles task-relevant from task-irrelevant information. Following prior work \cite{sutter2024unity}, multimodal performance for the SOTA methods is reported as the average of single-modality results; multimodal evaluation uses paired samples, while unimodal results use all available data.

\subsubsection{Ablation Study}
Ablation results (Table \ref{tab:ablation}) highlight the importance of disentangling task-specific (\textit{fusion, uni}) and instance-specific (\textit{inst}) latent. Removing disentanglement substantially degrades multimodal performance, reducing AUROC from $0.725$ to $0.646$, while omitting either unimodal or fused task-relevant components also leads to lower performance ($0.658$--$0.680$), indicating that both modality-specific and multimodal information are important for prediction.

\begin{table}[t]
\centering
\caption{\textbf{Ablation study of latent components.} Mean AUROC (95\% CI) for image-only, RNA-only, and multimodal prediction using linear probing, comparing a non-disentangled baseline and models with ablated task-relevant (\textit{uni}, \textit{fusion}) and instance-specific (\textit{inst}) latent components.}
\label{tab:ablation}
{\fontsize{8}{10}\selectfont
\begin{tabular}{ccccccc}
\toprule
\multicolumn{3}{c}{Ablation} &
IMG-only & RNA-only & Multimodal \\
$z^{uni}$ & $z^{fusion}$ & $z^{inst}$ & AUROC & AUROC & AUROC \\
\midrule

\xmark & \xmark & \xmark &
0.524 [0.488,0.559] &
0.643 [0.606,0.680] &
0.646 [0.611,0.682] \\

\cmark & \cmark & \xmark &
0.555 [0.505,0.606] &
\underline{0.674 [0.626,0.723]} &
0.680 [0.632,0.730] \\

\cmark & \xmark & \cmark &
\underline{0.603 [0.508,0.671]} &
\textbf{0.706 [0.669,0.748]} &
\underline{0.719 [0.693,0.760]} \\

\xmark & \cmark & \cmark &
0.524 [0.468,0.573] &
0.647 [0.604,0.681] &
0.658 [0.611,0.670] \\

\midrule

\cmark & \cmark & \cmark &
\textbf{0.638 [0.601,0.673]} &
\textbf{0.706 [0.666,0.741]} &
\textbf{0.725 [0.682,0.757]} \\

\bottomrule
\end{tabular}
}
\end{table}

\subsubsection{Analysis of Explainability}
\begin{figure}[!b]
\centering

\subfloat[Quantitative Results\label{tab:dice}]{
\begin{minipage}[c]{0.40\textwidth}
\centering
{\fontsize{8}{10}\selectfont

\begin{tabular}{lc}
\toprule
\multicolumn{2}{l}{\textit{\textbf{Prediction-based}}}\\
\midrule
Method & DICE \\
\midrule
ImgVAE & 0.043$\pm$0.023 \\
ContIG \cite{taleb2022contig} & \underline{0.080$\pm$0.003} \\
MMVM VAE \cite{sutter2024unity} & 0.007$\pm$0.002 \\
DMVAE \cite{lee2021private} & 0.016$\pm$0.004 \\
SYNPRED (ft) & \textbf{0.167$\pm$0.002} \\
\bottomrule
\end{tabular}

\par\smallskip
\begin{tabular}{ccc c}
\multicolumn{4}{l}{\textit{\textbf{Representation-based}}}
 \\
\midrule
$z^{uni}$ & $z^{fusion}$ & $z^{inst}$ & DICE
\\
\midrule
\xmark & \xmark & \xmark & 0.069$\pm$0.088 \\
\xmark & \cmark & \cmark & \underline{0.121$\pm$0.141} \\
\cmark & \cmark & \xmark & 0.113$\pm$0.016 \\
\cmark & \xmark & \cmark & 0.120$\pm$0.147 \\
\cmark & \cmark & \cmark & \textbf{0.133$\pm$0.128} \\
\bottomrule
\end{tabular}
}
\end{minipage}
}
\hfill
\subfloat[Qualitative Results\label{fig:saliency}]{
\begin{minipage}[c]{0.56\textwidth}
\centering
\includegraphics[width=\linewidth]{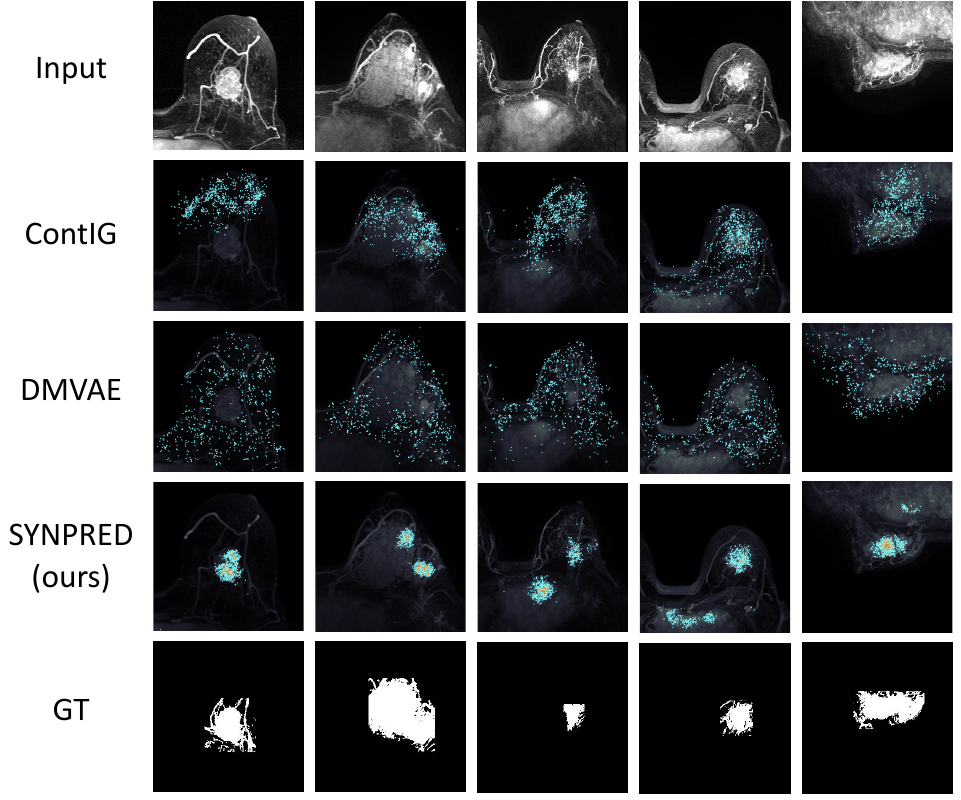}
\end{minipage}
}
\caption{\textbf{Explainability evaluation.}
(\textit{Left}) DICE overlap between saliency maps and ground-truth segmentations. (\textit{Right}) Qualitative saliency visualizations. The proposed
method achieves the highest overlap and more coherent localization of task-relevant regions.}
\end{figure}

The proposed method is able to improve localization of clinically relevant regions that are informative for the prediciton task. We assess model's focus on tumour region by comparing it to saliency-based explanations and assessing the overlap with ground-truth segmentations of lesions (Fig. \ref{tab:dice}). The proposed method achieves the highest overlap compared to SOTA approaches, reaching a DICE score of $0.167$. In representation-based evaluation, the full disentangled model consistently outperforms not only the non-disentangled variant but also all ablated configurations, indicating that jointly modeling modality-specific and multimodal task-relevant information yields the most spatially coherent explainability maps.

\section{Discussion}
Disentangling complementary and shared task-relevant information, rather than enforcing cross-modal alignment, is important for multimodal learning in settings with limited semantic overlap, such as radiology-genomics applications combining RNA and imaging data. This is particularly relevant when modalities have asymmetric predictive strength, with imaging acting as the weaker modality. Existing multimodal generative approaches for medical data \cite{taleb2022contig,ai2024sama,li2024mri} primarily emphasize shared information, often underutilizing modality-specific yet task-relevant signals. In contrast, our framework explicitly disentangles shared, modality-specific, and instance-specific representations to better capture complementary multimodal information.


Empirical results suggest that disentanglement and explicit modeling of modality-specific and multimodal components contribute to robust prediction and clinically meaningful localization. The proposed method achieves stronger performance than alignment-based and multimodal VAE baselines in multimodal prediction (Table \ref{tab:results}). Importantly, multimodal training also improves single-modality prediction, particularly for imaging, which has lower standalone predictive power and benefits most from complementary RNA information. These gains provide empirical evidence that synergistic cross-modal interactions during multimodal training can uncover predictive structure in the weaker imaging modality that is difficult to capture from imaging alone. Ablation studies (Table \ref{tab:ablation}) further indicate that both the \textit{uni} and \textit{fusion} components contribute to performance, while explainability analysis shows improved overlap with ground-truth segmentations compared to alignment-based and VAE baselines. While applicable in principle, the framework was not evaluated on settings with more than two modalities or on tasks beyond classification. In addition, the current evaluation is limited to a single radiology-genomics benchmark involving paired breast MRI and RNA data, and broader validation on additional datasets is needed to assess generalizability.

\section{Conclusion}
We present a multimodal deep generative framework for clinical prediction from medical modalities with limited semantic overlap. By explicitly disentangling modality-specific and shared task-relevant representations from instance-specific variation, the proposed approach improves representation learning from complementary multimodal signals. Empirical results show consistent gains in single-modality prediction after multimodal training, particularly for imaging with limited standalone predictive power, while maintaining competitive multimodal performance. These findings provide empirical evidence that synergistic cross-modal interactions during multimodal training can uncover predictive structure in the weaker imaging modality that would otherwise remain difficult to capture.

\begin{credits}
\subsubsection{\ackname} This work was funded by the Vienna Science and Technology Fund (WWTF, PREDICTOME [10.47379/LS20065]) the European Union’s Horizon Europe research and innovation programme under grant agreement No.101100633— EUCAIM, Austrian Federal Ministry of Labour and Economy, the National Foundation for Research, Technology and Development and the Christian Doppler Research Association, and Siemens Healthineers. 
\subsubsection{\discintname}The authors have no competing interests to declare that are relevant to the content of this article.
\end{credits}
%
%
%

%
\bibliographystyle{splncs04}
\bibliography{bibliography}

\end{document}